\documentclass[runningheads]{llncs}
\usepackage{graphicx}
\usepackage{multirow}
\usepackage{amsmath}
\usepackage{amsfonts}
\usepackage{cite}
\usepackage{comment}
\usepackage{tabularx}
\begin{document}

\title{Cell Detection in Domain Shift Problem Using Pseudo-Cell-Position Heatmap}

\author{
Hyeonwoo Cho\inst{1} \and Kazuya Nishimura\inst{1} \and Kazuhide Watanabe\inst{2} \and Ryoma Bise\inst{1}
}
\authorrunning{Hyeonwoo Cho et al.}
\titlerunning{Cell Detection in Domain Shift Problem Using Pseudo Heatmap}
\institute{Kyushu University, Fukuoka City, Japan.
\email{hyeonwoo.cho@human.ait.kyushu-u.ac.jp}
\and RIKEN Center for Integrative Medical Sciences, Japan}

\maketitle 

\begin{abstract}
The domain shift problem is an important issue in automatic cell detection. A detection network trained with training data under a specific condition (source domain) may not work well in data under other conditions (target domain). We propose an unsupervised domain adaptation method for cell detection using the pseudo-cell-position heatmap, where a cell centroid becomes a peak with a Gaussian distribution in the map. In the prediction result for the target domain, even if a peak location is correct, the signal distribution around the peak often has a non-Gaussian shape. The pseudo-cell-position heatmap is re-generated using the peak positions in the predicted heatmap to have a clear Gaussian shape. Our method selects confident pseudo-cell-position heatmaps using a Bayesian network and adds them to the training data in the next iteration. The method can incrementally extend the domain from the source domain to the target domain in a semi-supervised manner. In the experiments using 8 combinations of domains, the proposed method outperformed the existing domain adaptation methods.
\end{abstract}

\section{Introduction}
Cell detection has an important role in quantification in bio-medical research. Automatic cell-detection methods can be classified as traditional image-processing-based methods and deep-learning-based methods. Image-processing-based methods are generally designed based on the image characteristics of the target images. For example, methods using thresholding~\cite{otsu1979threshold, yuan2012quantitative}, image filters~\cite{cosatto2008grading}, region growing~\cite{zhou2004segmentation}, watershed~\cite{yuan2012quantitative}, and graph cuts~\cite{al2009improved} use the intensity or edge information. These methods work under the specific conditions used for developing them but often not under others.

Deep-learning-based methods have recently achieved very promising results in cell-detection problems \cite{yi2019multi, li2019signet, kainz2015you, nishimura2019weakly, fujita2020cell}. However, a network trained with training data under a specific condition (source domain), {\it e.g.}, culture conditions, may not work well in data under other conditions (target domain) since the image features differ among different domains (domain shift problem). For example in Fig. 1, although both the source and target domains are the same cell type, the shapes of cells in the target domain cultured under different conditions are elongated compared with those in the source domain (the F-score decreased from 0.941 (source) to 0.768 (target) in our experiments). This indicates that it is necessary to prepare training data for not only cell types but also the different culture conditions, even though the condition often changes depending on the purpose of biological research.

\begin{figure}[t!]
 \centering
 \includegraphics[width=0.95\textwidth]{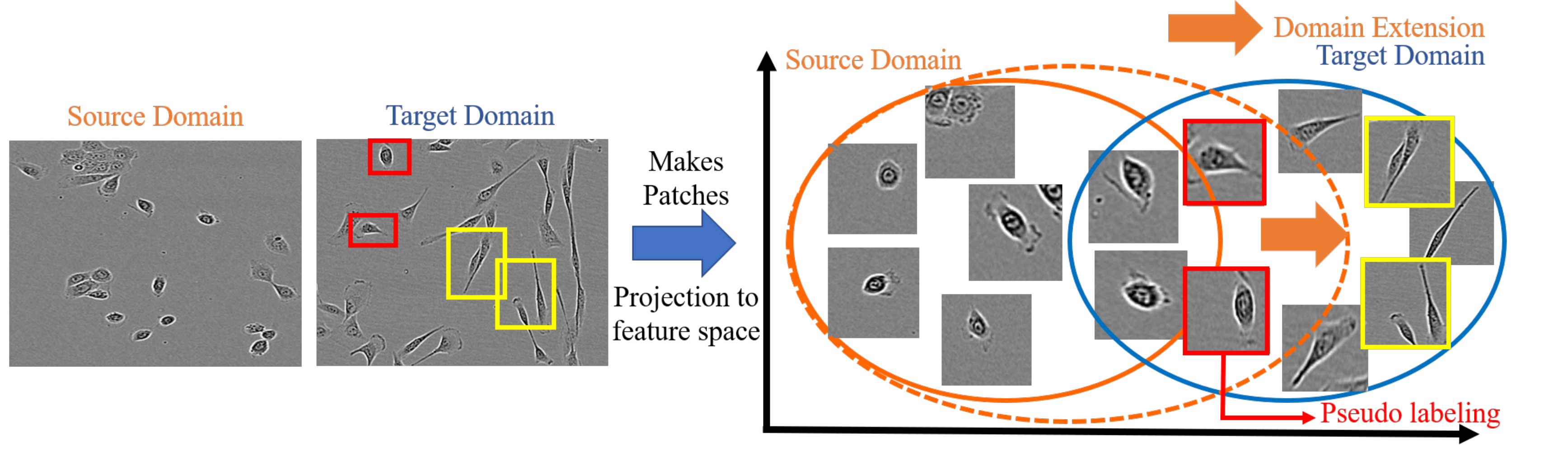}
 \caption{\textbf{Left} shows examples of cell images on source and target domain and red and yellow rectangles show cells in target with similar and different shape to those in source, respectively. \textbf{Right} shows illustration of feature space of patches cut from left images, where orange arrow indicates domain extension.}\label{fig1}
\end{figure}

To address the domain shift problem, unsupervised domain adaptation (UDA) methods, which use only data on the source domain without data on the target domain, have been proposed \cite{ghifary2016deep,ganin2015unsupervised, tzeng2017adversarial, ge2020domain, jin2018unsupervised, saito2019strong, haq2020adversarial, tsai2018domain}. Most are designed for classification tasks but not for detection tasks. A major approach of UDA method uses adversarial learning that transfers the distribution of the target to the source's in the same feature \cite{tzeng2017adversarial, ganin2015unsupervised, cui2020gradually, haq2020adversarial}. Haq {\it et al.} \cite{haq2020adversarial} extended this to the cell-segmentation task by introducing an auto-encoder. This method transfers the feature distribution of an entire image, in which it is implicitly assumed that the characteristics, such as illumination, and color, of the entire image differ between the source and target domains.
However, our target has a deference difficulty from entire image adaptation, where an image contains many cells with various appearances ({\it e.g.}, shapes, density), as shown in the red and yellow boxes in Fig. \ref{fig1}.
It is important to consider the different appearances of cells in an image for further improvement.
Another type of domain adaptation method uses semi-supervised learning for selecting pseudo labels from data in the target domain, and the network is re-trained using those pseudo labels \cite{kim2020attract,li2020online,jiangbidirectional, zheng2020rectifying,saito2019semi, ge2020mutual,wang2020unsupervised}, in which our method is categorized to this approach. The main idea of pseudo labeling is that some samples in the target domain can be correctly classified, and if we can select correctly predicted samples as the additional training data, the performance improves. However, such methods were designed for classification tasks.

In this paper, we propose a cell-detection method that addresses the domain shift problem and also improves detection performance in the same domain in a semi-supervised manner. To handle various cell appearances in an image, our method first separates an image into patches. Some patch images in the target are often similar to some in the source, {\it i.e.}, the image feature distributions between domains partially overlap, as shown in Fig. \ref{fig1}, which is assumed in most UDA methods. 
In our preliminary study, we used the cell-position-heatmap prediction method \cite{cao2019openpose, nishimura2019weakly}, which is one of the state-of-the-arts, where a cell centroid becomes a peak with a Gaussian distribution in the map.
The key observation is that it could detect a cell that has a similar but slightly different shape from the source's even the prediction map is a non-Gaussian shape as shown in Fig. \ref{fig2}.
Moreover, the detection performance improved using the correctly detected cells with clear Gaussian maps (pseudo-cell-position heatmaps) as training data in our preliminary study. Since it is important to select the confident patches as pseudo labels, we introduce a Bayesian discriminator that can estimate the uncertainty for each patch. We then use the selected patches with the clear Gaussian maps as pseudo-cell-position heatmaps for re-training the detector. These processes are iteratively performed. Since this process incrementally adds the confident pseudo labels, as shown in Fig. \ref{fig1} (called domain extension), this can improve detection performance both on the source and target domains. The main contributions of this paper are as follows:
\begin{itemize}
 \item We propose the unsupervised domain adaptation method for cell detection using the pseudo-cell-position heatmap that can incrementally extend the domain from the source to the target in a semi-supervised manner.
 \item We introduce a Bayesian discriminator that can estimate the uncertainty of each patch for selecting confident pseudo labels with high certainty under a self-training framework.
 \item The proposed method is applicable not only to the same domain but also to different domains with a small amount of training data. We confirmed the effectiveness of the method regarding detection performance through experiments involving 8 combinations of domains.
\end{itemize}

\section{Domain extension in cell detection}
\begin{figure}
    \centering
    \includegraphics[width=0.81\textwidth]{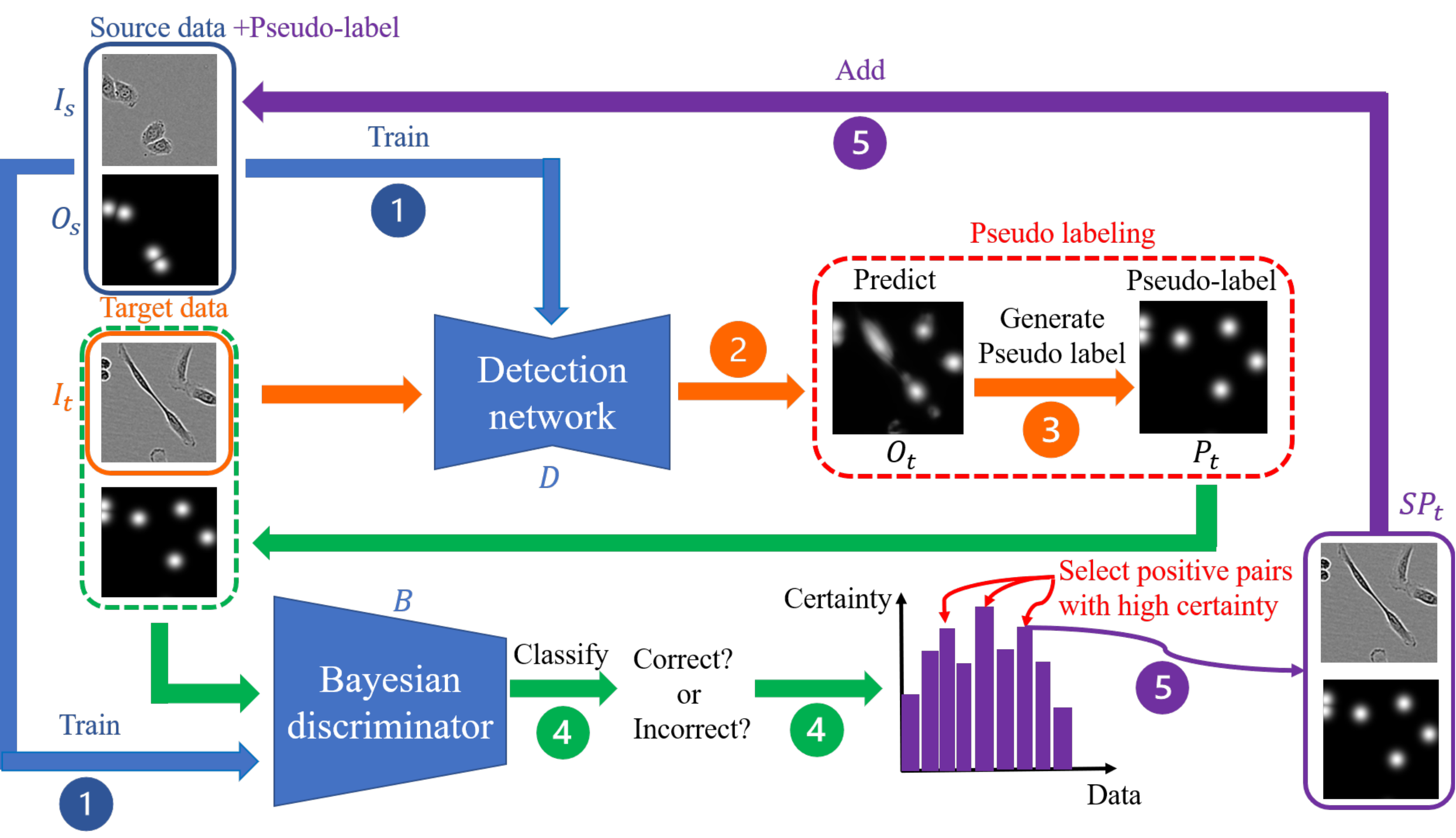}
    \caption{Overview of proposed method. Blue arrows represent training detection network and Bayesian discriminator with source data and pseudo labels. Orange arrows represent predicting heatmap of unlabeled data on target domain and generating pseudo-cell-position heatmap. Green and purple arrows represent selecting positive pseudo-cell-position heatmaps with high certainty by Bayesian discriminator, and adding selected patches as pseudo heatmap. This flow is iteratively performed.}
    \label{fig2}
\end{figure}

Fig. \ref{fig2} shows an overview of our method. Initially, the entire images are separated to patch images that may contain several cells (1 to 10 cells). We denote $I_{s}$, $O_{s}$ as the set of the original patch images and the ground truth of the cell-position heatmap on the source domain, and $I_{t}$ as the unlabeled images on the target domain. The proposed method consists of five steps. In step 1, using $I_{s}$ and $O_{s}$, we train the detection network $D$ that estimates the cell-position heatmap \cite{nishimura2019weakly} and the Bayesian discriminator $B$ that estimates the uncertainty of the estimated cell-position heatmap. In step 2, $D$ estimates this heatmap for $I_{t}$, in which the predicted cell-position heatmap $O_{t}$ may have a distorted shape (non-Gaussian) even if the peak position is correct, as shown in Fig. \ref{fig2}. In step 3, the pseudo cell position heatmaps (pseudo-ground-truth) $P_{t}$ with clear Gaussian shapes are generated on the basis of $O_{t}$. In step 4, given pairs of the original $I_{t}$ and its $P_{t}$ as inputs, $B$ estimates the uncertainty score if the inputted pseudo label is correct. In step 5, the method selects confident pseudo-position heatmaps $SP_{t}$ with high certainty, which are used as the additional training data for $D$ and $B$. This process is iteratively performed on $I_{t}$.

\noindent
\textbf{Cell detection with cell-position heatmap (Steps 1 and 2):}
Object-detection tasks often use a bounding box as the ground truth for localizing objects. However, the bounding box is not suitable for cell detection due to its complex shape and high density since it may often contain other cells. Instead, we use a cell-position heatmap that has produced a good performance \cite{nishimura2019weakly}. Given the set of annotated cell centroid positions for an image, a ground truth of the cell-position heatmap is generated so that a cell centroid becomes a peak with a Gaussian distribution in the map, as shown in Fig. \ref{fig2}. To train $D$, we use the mean squared error (MSE) loss function between the estimated image and ground-truth heatmap. 

\noindent
\textbf{Pseudo labeling for cell detection (Step 3):}
The predicted cell-position heatmap $O_{t}$ in the target domain may have a distorted Gaussian distribution, even if the peak position is correct. We generate the pseudo-cell-position heatmaps $P_t$ on the basis of $O_{t}$ so that detected positions $C_{t}$ in the predicted cell-position heatmap becomes a peak with a clear Gaussian distribution in the same manner as in step 1. Here, if the peak of $O_{t}$ is higher than a threshold $th_d$, the position $C_{t}$ is detected. Examples of predicted and pseudo-cell-position heatmaps are shown in the red dotted box in Fig. \ref{fig2}. The pseudo-cell-position heatmap will be used for training $D$ and $B$. Even if the signals of a cell in the initially estimated cell-position heatmap have a non-Gaussian shape, cells that have a similar appearance to the pseudo labels can be detected with a clear Gaussian shape by the re-trained $D$ using the pseudo-cell-position heatmap. The $D$ will incrementally improve from the iteration of this pseudo-labeling process.

\noindent
\textbf{Bayesian discriminator (Steps 4 and 5):}
If a pseudo-cell-position heatmap contains many incorrect labels, detection performance may be affected. To select confident pseudo-cell-position heatmaps $SP_t$, we introduce a Bayesian CNN to $D$ that estimates uncertainty in whether the estimated detection result is correct. To represent model uncertainty, we leverage a dropout-based approximate uncertainty inference \cite{gal2015bayesian, gal2016dropout}, in which the variance of the estimation of samples from the posterior distribution can be considered as an uncertainty measure.

The model with dropout learns the distribution of weights at training. We then sample the posterior weights of the model by using the dropout at test to find the predictive distribution over output from the model \cite{gal2015bayesian, gal2016dropout}. In practice, at test, the averaging stochastic forward passes through the model with dropout and averaging results are identical to the variance to model uncertainty\cite{gal2015bayesian, gal2016dropout}. We use this as a measure of model uncertainty. Model uncertainty is defined as
\noindent 
\begin{equation}
Uncertainty \approx \frac{1}{T}\sum_{t=1}^{T}p(\,\widehat{y}^{\,\ast}\mid\widehat{x}^{\,\ast},\widehat{w}_{t}),
\end{equation}
where $\widehat{w}_{t}$ are weights sampled from the distribution over the model's weights and $T$ is the number of sampling (the number of networks with dropout), $\widehat{x}^{\,\ast}$ and $\widehat{y}^{\,\ast}$ are input and output from the model. This model is referred to as MC dropout \cite{gal2016dropout}.

In our model, the input of the network in inference is a pair of the original image and generated pseudo-cell-position heatmap $\mathbf{X}^{(k)}=\{I_t^{(k)}, P_t^{(k)}\}$, and the output is the label $Y=\{0,1\}$ with $Uncertainty$, where $k$ is the data index. If $P_t^{(k)}$ looks like a correct label for the original image $I_t^{(k)}$, $Y$ takes 0; otherwise, 1. If the estimated label is confident, it produces lower uncertainty; otherwise, higher uncertainty. To train $B$ in the initial iteration, we deliberately make the incorrect ground truth from the ground truth of $I_{s}$ as negative samples by adding or removing and shifting the Gaussian in the ground truth of $I_{s}$. In inference, we $T$ times apply the network for a single input with the different dropout and obtain the predicted labels with uncertainties for all image pairs of $\{I_t, P_t\}$ based on the $T$ results. We then select $SP_{t}$ that have with $th_u$ lowest uncertainty and add them as training data for $D$ and $B$ in the next step.

\section{Experiment}
We evaluated the effectiveness of our method by using 8 combinations of domains. The baseline (Sup.) of the proposed method is a supervised method \cite{nishimura2019weakly} that uses training data on the source only. We also compared the proposed method with four other methods; an unsupervised image-processing-based method proposed by Vicar \cite{vicar2019cell} that combined Yin \cite{yin2012understanding} and distance transform, a semi-supervised learning method proposed by Moskvyak \cite{moskvyak2021semi} using consistency of prediction and is not for domain adaptation, a method (Haq w/o AE) \cite{haq2020adversarial} that simply introduces adversarial domain adaptation for cell segmentation, and a domain adaptation method proposed by Haq \cite{haq2020adversarial} that introduced auto-encoder to adversarial learning for cell segmentation.
\begin{table}[t]
\begin{center}
\caption{Performance of proposed methods on same and target domains}
\label{tab 3:performace on different domain}
\begin{tabular}{lc|cc|cccccccc}
\hline 
Data & S to T & \begin{tabular}{c}Sup.\\ on S\end{tabular} & \begin{tabular}{c}Ours\\ on S\end{tabular} & \begin{tabular}{c}Sup.\\ on T\end{tabular} & \begin{tabular}{c}{\itshape Vicar} \\ \cite{vicar2019cell}\end{tabular} & \begin{tabular}{c}{\itshape Moskvyak} \\ \cite{moskvyak2021semi}\end{tabular} & \begin{tabular}{c}{\itshape Haq w/o AE} \\ \cite{haq2020adversarial}\end{tabular} & \begin{tabular}{c}{\itshape Haq} \\ {\cite{haq2020adversarial}}\end{tabular} & \begin{tabular}{c}Ours\\ on T\end{tabular} \\
\hline\hline
\multirow{6}{*}{C2C12} & F → C & 0.800 & \textbf{0.833} & 0.684 & 0.700 &0.642 &0.766 &0.771 & \textbf{0.845} \\
                       & B → C & 0.885 & \textbf{0.941} & 0.756 & 0.700 &0.775 &0.843 &0.848 & \textbf{0.860} \\
                       & C → F & 0.850 & \textbf{0.867} & 0.705 & 0.612 &0.642 &0.755 &0.767 & \textbf{0.807} \\
                       & B → F & 0.885 & \textbf{0.899} & 0.632 & 0.612 &0.648 &0.766 &0.771 & \textbf{0.798} \\
                       & C → B & 0.850 & \textbf{0.862} & 0.709 & 0.584 &0.761 &0.779 &0.795 & \textbf{0.820} \\
                       & F → B & 0.800 & \textbf{0.835} & 0.742 & 0.584 &0.721 &0.761 &0.757 & \textbf{0.817} \\
\hline
\multirow{2}{*}{HMEC} & C → E & 0.941 & \textbf{0.962} & 0.768 & 0.797 &0.798 &0.851 &0.849 & \textbf{0.875}\\
                        & E → C & 0.941 & \textbf{0.971} & 0.939 & 0.857 &0.835 &0.924 &0.921 & \textbf{0.959}\\
\hline\hline
\multirow{1}{*}{} &Average  &0.869 & \textbf{0.896} &0.742 & 0.681 &0.728 &0.806 &0.810 &\textbf{0.848} \\
                    
\hline\hline
\end{tabular}
\end{center}
\end{table}

\begin{table}[t!]
      \centering
        \caption{Performance when proposed method was used as semi-supervised learning using additional data in same domain.}
        \label{tab 1:Performance on same domain}
        \begin{tabular}{lccccc}
        \hline 
        Data & Cond. & Sup. & {\itshape Vicar \cite{vicar2019cell}}& {\itshape Moskvyak\cite{moskvyak2021semi}} &Ours\\
        \hline\hline
        \multirow{3}{*}{C2C12} & C & 0.850 &0.700&0.851 & \textbf{0.888} \\
                               & B & 0.885 & 0.584 &0.899 & \textbf{0.924} \\
                               & F & 0.800 & 0.612 &\textbf{0.859} & 0.842 \\
        \hline
        \multirow{2}{*}{HMEC} & C & 0.941 & 0.857 &\textbf{0.963} & 0.961\\
                                & E & 0.941 &0.797 &0.958 & \textbf{0.961}\\
        \hline
        \multirow{1}{*}{} &Average &0.883 &0.710 &0.906 &\textbf{0.915} \\
        \hline\hline
        \end{tabular}
\end{table}

\noindent
\textbf{Implementation details:}
In this experiment, We employed U-Net as the detection network and Resnet18 as the discriminator, and the four iterations (Iters 0 to 3) were performed in all experiments. We used Adam to optimize our model and set the learning rate is 1.0 $\times$ $10^{-3}$ and all datasets are normalized between 0 and 255. Hyperparameters to the proposed method are $th_d$, $th_u$ and $T$ which are 100, 10 and 10 in all experiments.

\begin{figure}[t!]
    \centering
    \includegraphics[width=0.7\textwidth]{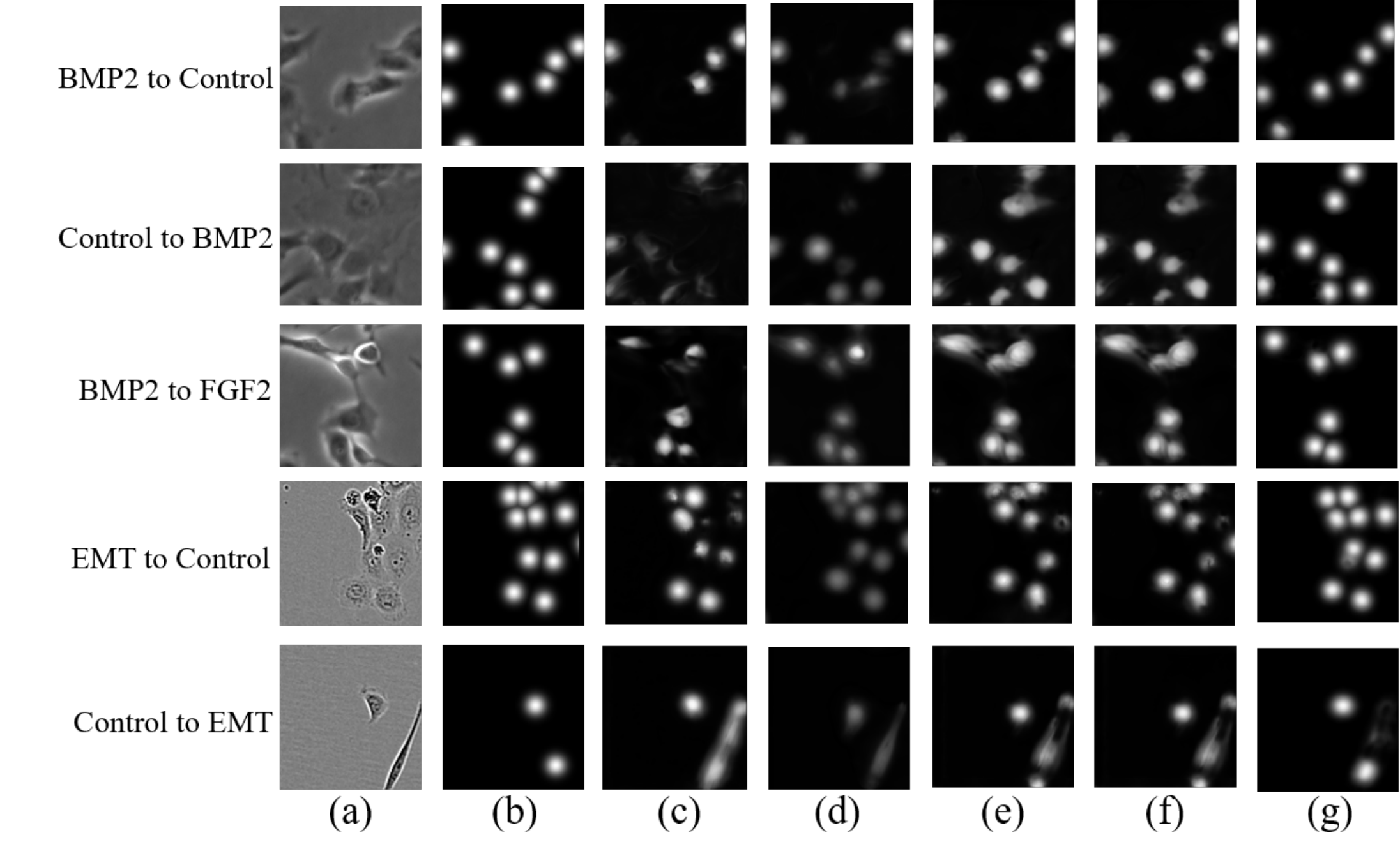}
    \caption{(a) Original image, (b) Ground-truth, (c)Sup., (d)Moskvyak, (e) Haq w/o AE, (f) Haq, (g) Ours} \label{fig4: Estimation_result}
\end{figure}

\noindent
\textbf{Datasets and evaluation metric:}
To determine how our method performs regarding domain shift, we evaluated two datasets. The first is C2C12 \cite{eom2018phase} consisting of myoblast cells captured by phase contrast microscopy at a resolution of $1040 \times 1392$ pixels and cultured under three different conditions; 1) Control (no growth factor), 2) FGF2 (fibroblast growth factor), 3) BMP2 (bone morphogenetic protein). The second dataset is Human mammary epithelial cells (HMEC), consisting of cells captured by phase-contrast microscopy at a resolution of $1272 \times 952$ and cultured under two conditions; 1) Control (no stimulus) and 2) EMT (epithelial-mesenchymal transition) \cite{nieto2016emt}. As shown in Fig. \ref{fig4: Estimation_result} (a), the cell appearance under each condition is different. We made patches of $128 \times 128$ pixels from full images of all datasets, and the ground truth was given for only 24 patches (approximately a quarter of one entire image) under each condition. In the evaluation of domain adaptation, we used a sequence as unlabeled data (100 entire images), and the other sequence as test data (100 entire images) under different conditions from the training data.
We used F-score as the performance metric, which is measured by solving a one-by-one matching based on proximity of the estimated cell positions and the ground truth. The cell location is defined as the peak position of each Gaussian distribution. If the detected result is close enough to the given ground truth, it is counted as True Positive. When there is no ground truth around the detected location, it is counted as False Positive, and when there is no detected result around the ground-truth location, it is counted as False Negative. 

\noindent
\textbf{Evaluation:} 
Tab. \ref{tab 3:performace on different domain} shows the detection performances (F-score) in the target domain and source domain. The source and target are denoted as S and T and each culture condition (Cond.) is denoted by the first letter of the condition name. 'Sup. on S' indicates the performance of the baseline method when evaluated on the other sequences in the same domain. The performances of the baseline (Sup. on T) significantly decreased from those (Sup. on S) evaluated on the same domain (average: -0.127). Although Haq w/o AE and Haq's methods improved in performance in the target domain, our method outperformed these methods for all combinations. The columns (Sup. on S), (Ours. on S) in Tab. \ref{tab 3:performace on different domain} shows the performances in the additional test data on the source domain after applying our method. In all data, our method also improved in performance not only in the target domain but also in the source domain. We consider that the pseudo-cell-position heatmaps for cells that have similar appearances to those in the source domain improved the detection performance of the proposed method. Fig. \ref{fig4: Estimation_result} shows example detection results of the test data on the target domains. As shown in Fig. \ref{fig4: Estimation_result}(c), the supervised method did not effectively predict cell-position heatmaps, but the peaks in some were correctly detected. In contrast, our method could predict these maps with a clear Gaussian shape. These results support the idea of the proposed method.

Next, we evaluated the effectiveness of the proposed method as semi-supervised learning to confirm that it can improve detection performance from the few labeled and many unlabeled data. In this experiment, 24 patch images were used as labeled data, and 8000 patch images (100 entire images) were used as unlabeled data for each condition. The evaluation was then conducted in which all the images were in the same domain. Since Haq w/o AE and Haq's methods are not for semi-supervised learning, we did not evaluate them. Tab. \ref{tab 1:Performance on same domain} shows that our method performed better for all domains compared with the baseline, and the average was slightly better than Moskvyak's method \cite{moskvyak2021semi}, which was developed for semi-supervised learning.

\begin{figure}[t!]
 \centering
 \includegraphics[width=0.9\linewidth]{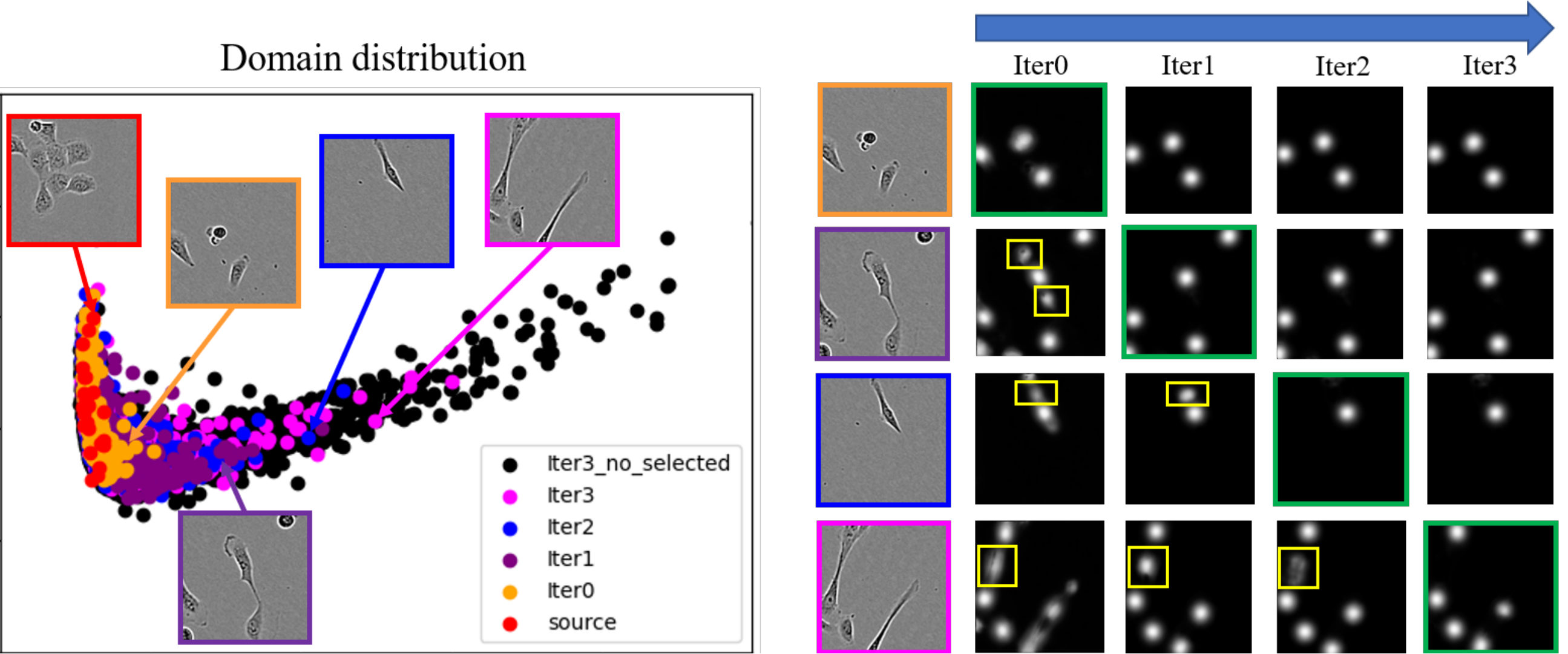}
 \caption{\textbf{Left} is feature distribution of source and target domains (Iteration 0 to 3-no-selected). \textbf{Right} represents cell shape and detection result at each iteration. Green prediction cell-position map signifies correct detection results and yellow boxes represent False Positive.} \label{fig5: Domain Distribution}
\end{figure}

\noindent
\textbf{Visualization of domain-extension process:}
The left image in Fig. \ref{fig5: Domain Distribution} shows the feature distribution of the source (control of HMEC) and the target domain (EMT), where each iteration is marked with a different color. In this map, the cells in the source domain are round (red) and distributed on the left side, and the cells in the target domain have various shapes (orange, purple, blue, magenta, and black) and distributed from the left to right. The colored box images are the samples from this distribution. We can observe that the proposed method incrementally increased the pseudo patches from left to right in the feature map with iteration. The right image in Fig. \ref{fig5: Domain Distribution} shows examples of estimated cell-position heatmaps in each iteration. In iteration 0, the elongated cell, which is located on the right side of the feature map, was mistakenly detected as several cells. Such cells far from the source images in the map were not selected in the early iterations, but the prediction results of the heatmap improved in the later iterations, which were finally selected when the prediction was successful. These results indicate that the discriminator performed well in finding the correct pseudo-position heatmap, and the proposed method incrementally extended the domains from the source to target.

\section{Conclusion}
We proposed a domain adaptation method for cell detection based on semi-supervised learning by selecting pseudo-cell-position heatmaps using model uncertainty. The experiment results using various combinations of domains demonstrated the effectiveness of our method, which improved the performance of detection on unannotated cells on not only different domains but also source domains. Also, the analysis of each iteration demonstrated that the method incrementally extended the domain from the source and target.

\noindent
{\bf Acknowledgment:}
This work was supported by JSPS KAKENHI Grant Number JP20H04211 and JP21K19829.

\bibliographystyle{splncs04}
\bibliography{refer}

\end{document}